\documentclass[10pt,twocolumn,letterpaper]{article}

\usepackage[pagenumbers]{cvpr} % To force page numbers, e.g. for an arXiv version

\usepackage{graphicx}
\usepackage{amsmath}
\usepackage{amssymb}
\usepackage{booktabs}
\usepackage{times}
\usepackage{epsfig}
\usepackage{mathtools} 
\usepackage{algorithm}
\usepackage{algorithmic}
\usepackage{threeparttable}
\usepackage{multirow} 
\usepackage{comment}

\usepackage[pagebackref,breaklinks,colorlinks]{hyperref}

\usepackage[capitalize]{cleveref}
\crefname{section}{Sec.}{Secs.}
\Crefname{section}{Section}{Sections}
\Crefname{table}{Table}{Tables}
\crefname{table}{Tab.}{Tabs.}

\def\cvprPaperID{0} % *** Enter the CVPR Paper ID here
\def\confName{CVPR}
\def\confYear{2023}

\begin{document}
	
	% \title{\LaTeX\ Author Guidelines for \confName~Proceedings}
	\title{Discovering Local Binary Pattern Equation for Foreground Object Removal in Videos}
	
	% 	Institution1\\
	% 	Institution1 address\\
	% 	{\tt\small firstauthor@i1.org}
	% 	% For a paper whose authors are all at the same institution,
	% 	% omit the following lines up until the closing ``}''.
	% 	% Additional authors and addresses can be added with ``\and'',
	% 	% just like the second author.
	% 	% To save space, use either the email address or home page, not both
	% 	\and
	% 	Second Author\\
	% 	Institution2\\
	% 	First line of institution2 address\\
	% 	{\tt\small secondauthor@i2.org}
	% }
	\author{Caroline Pacheco do Espirito Silva\\
		Machine Learning Team, ActiveEon\\
		Paris, France\\
		{\tt\small lolyne.pacheco@gmail.com}
		\and
		Andrews Cordolino Sobral\\
		Machine Learning Team, ActiveEon\\
		Paris, France\\
		{\tt\small andrewssobral@gmail.com}
		\and
		Antoine Vacavant\\
		CNRS, SIGMA Clermont, Institut Pascal\\
        University of Clermont Auvergne\\
		Clermont-Ferrand, France\\
		{\tt\small antoine.vacavant@udamail.fr}
		\and
		Thierry Bouwmans\\
		Lab. MIA, University La Rochelle\\
        University of La Rochelle\\
		La Rochelle, France\\
		{\tt\small thierry.bouwmans@univ-lr.fr}
		\and
		Felippe De Souza\\
		Department of Electromechanical Engineering\\
        University of Beira Interior\\
		Covilha, Portugal\\
		{\tt\small felippe@ubi.pt}
	}

	\maketitle
	
	%%%%%%%%% ABSTRACT
	\begin{abstract}
Designing a novel Local Binary Pattern (LBP) process usually relies heavily on human experts' knowledge and experience in the area. Even experts are often left with tedious episodes of trial and error until they identify an optimal LBP for a particular dataset. To address this problem, we present a novel symbolic regression able to automatically discover LBP formulas to remove the moving parts of a scene by segmenting it into a background and a foreground. Experimental results conducted on real videos of outdoor urban scenes under various conditions show that the LBPs discovered by the proposed approach significantly outperform the previous state-of-the-art LBP descriptors both qualitatively and quantitatively. Our source code and data will be available online.
	\end{abstract}
	
	%%%%%%%%% BODY TEXT
	\section{Introduction}
	\label{sec:intro}
	
Background subtraction (BS) is an attractive research field in computer vision and video processing. It has received increasing attention over the last few decades and has remained a very active research direction, thanks to the numerous potential applications and the availability of surveillance cameras installed in security-sensitive areas such as banks, train stations, highways, and borders ~\cite{jain:2018, sobral:2014}. BS aims to obtain an effective and efficient background model to remove the moving parts of a scene by segmenting it into background and foreground  ~\cite{garcia:2020}. Generally, it is challenging to design a promising BS algorithm in real environments due to the sudden illumination changes, dynamic backgrounds, bad weather, noise, and strong shadows. Several visual feature representations have been proposed to deal with these situations. Color intensities are the classic features used in BS, but they only reflect the visual perception properties of scene pixels, and usually discard the spatial information between adjacent pixels, resulting in the sensitivity to noise and sudden illumination changes.
	\begin{figure}[ht!]
		\centering
		\includegraphics[width=0.47\textwidth]{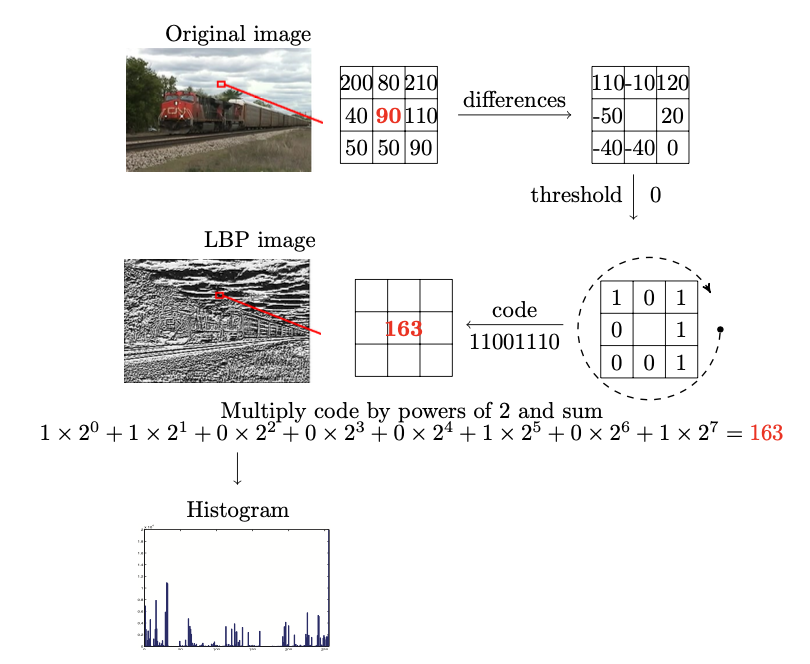}
		\caption{The LBP descriptor. From the original image to the histogram of its LBP image.}
		\label{chap2:fig:lbp}
	\end{figure}
	
	\begin{figure*}[t]
		\centering
		\scalebox{0.92}{
			\label{fig:framework1}
			\centering
			\includegraphics[width=0.9\textwidth]{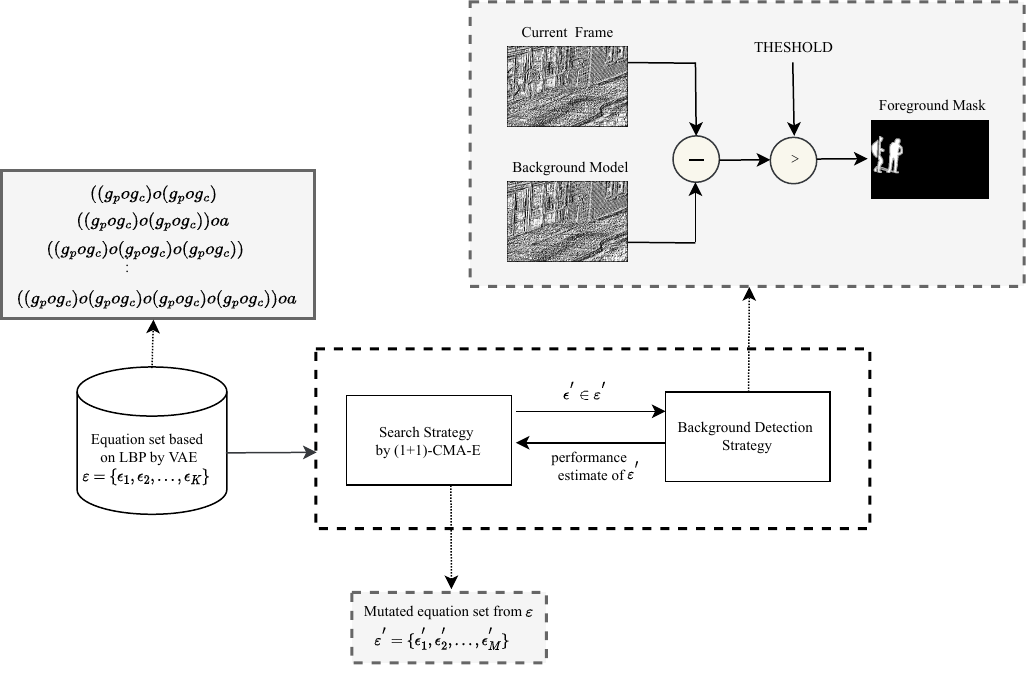}}
		\caption{\small Brief overview of the proposed framework. A set of equations based on LBP $\varepsilon = \{\epsilon_1,  \epsilon_2, \dots, \epsilon_K \}$, where $\varepsilon$ expresses each equation and $k$ is a user parameter that determines the number of elements $\epsilon \subseteq \varepsilon$. Initially, the (1+1)-CMA-ES seeks the best equation by mutating the arithmetic operators of each equation $\epsilon \subseteq \varepsilon$ resulting in a new one of the mutated equations $\varepsilon^{'} = \{\epsilon^{'}_1,  \epsilon^{'}_2, \dots,  \epsilon^{'}_M \}$. The performance of each equation is estimated by a background subtraction algorithm to distinguish the moving objects from the background of a set of videos. Finally, the $\epsilon^{'}$ that presented the maximum accuracy is selected as a best equation.}
		\centering
	\end{figure*}
	
A variety of local texture descriptors have attracted great attention for BS, especially the Local Binary Pattern (LBP)\cite{ojala:2002} because it is simple and quick to calculate. It is a powerful gray scale invariant texture descriptor. The computation of the LBP for a neighborhood of size $P$ = 8 is illustrated in Figure \ref{chap2:fig:lbp}. It combines the characteristics of statistical and structural texture analysis, describing the texture with micro-primitives and  their statistical placement rules. The LBP presents many properties which allow its use in BS, especially because LBP shows great invariance to changes in monotonic lighting common in natural images. It does not require many parameters to be set, and has a high discriminative power. Nevertheless, the original LBP formula has been reformulated in recent years to make it capable of dealing with several challenges found in different types of scenes, such as dynamic backgrounds, bad weather, noise, shadows, and others ~\cite{bouwmans:2018}. Discovering effective hand-crafted formulas based on LBP is not an easy task as it requires a deep knowledge of the scene and a trial and error process by experts until they identify a LBP formula that achieves meaningful results for a particular dataset. Moreover, human design cannot fully explore the space of all possible mathematical LBP formulas, often resulting in sub-optimal LBPs. However, we believe that it is possible to automatically discover accurate and efficient LBP formulas. Although symbolic regression approaches are less used in BS,  they can improve the foreground segmentation in complex scenes thanks to their capability to discover a function 
$f : \mathbb{R}^\rho \to \mathbb{R}$ that represents a given dataset $(\mathcal{X}, \mathcal{Y})$,  where each input points $\mathcal{X}_i \in \mathbb{R}^\rho$, output points $\mathcal{Y}_i \in \mathbb{R}$ and $f$ is a symbolic mathematical equation~\cite{petersen:2021, sciadviu:2020}. In recent years, symbolic regression has reached a remarkable increase in its popularity ~\cite{schmidt:2009, kusner:2017, brunton:2016, sahoo:2018, hein:2018, kong:2019, sciadviu:2020, landajuela:2021, biggio:2021}. One crucial aspect for this progress is decades of machine learning research into several aspects of the field, ranging from search strategies to novel neural architectures ~\cite{feurer:2015, feurer:2019}. Symbolic regression can be reformulated as an optimization problem with an optimality criterion within a given search space. Unlike previous approaches that apply optimization algorithms to constrained search spaces heavily reliant on human design without considering its innovation potential, we propose a novel symbolic regression for foreground object removal that discovers LBP formulas that we have not yet imagined.
The proposed method is designed to automatically find the best LBP formula to distinguish the moving objects from a set of videos without requiring huge manual efforts of human experts. In our method, the search for the optimal solution is performed by a variant of Covariance Matrix Adaptation Evolution Strategy (CMA-ES)~\cite{igel:2007, hansen:2001} which is called  (1+1)-CMA-ES~\cite{igel:2006}, that seeks for the optimal LBP on the search space designed by a Variational Autoencoder (VAE)~\cite{kingma:2014}. In summary, our contributions are as follows:

	\begin{itemize}
		\item  a set of optimal features based on LBP for dealing with specific challenges (e.g., changes in lighting, dynamic backgrounds, noise, strong shadows and others) in a real-world scene.
		%a set of optimal formulas based on LBP that we may never have thought of yet for dealing with specific challenges (e.g., changes in lighting, dynamic backgrounds, noise, shadows and others) in a real-world scene. % a search space designed by VAE containing a wide variety of formulas based on LBP that we may never have thought of yet;
		%\item  an efficient symbolic regression capable of automatically finding new LBP that achieve outperforming when compared to the previous state-of-the-art LBP human-designed.
		%surpass the original LBP   to deal with real-world scenarios.
		%dataset~\citep{wang:2014}
		\item an efficient symbolic regression capable of automatically finding new LBP that outperforms the previous state-of-the-art human-designed LBP. 
		
		\item detailed results showing the potential of our approach to distinguish moving objects from the background in video sequences.
	\end{itemize}

	\begin{algorithm}[t]
		\caption{Equation Discovery}
		\label{alg:structure_discovery}
		%	\tiny
		% \small
		\normalsize
		\begin{algorithmic}[1]
			%\textit{$\delta(x)$}, number of base classifier $M$, user defined parameter %\textit{$\varepsilon$}
			\STATE \textbf{Require:} training set images $X$, small input LBP equation set given by user $D$, VAE instance $\nu$, (1+1)-CMA-ES instance, Texture-BGS instance, arithmetic operators  $A = \{+ ,  -, *, /\}$, user parameter $K$
			\STATE // generate a new unseen set of LBP equations 
			\STATE // $\varepsilon = \{\epsilon_1,  \epsilon_2, \dots, \epsilon_K \}$ 
			\STATE  $\{\varepsilon \} \leftarrow$ \textit{VAE (D)} 
			\STATE \textit{k} $\leftarrow 1$
			\STATE \textbf{repeat}
			\FOR{$k = 1:K$}
			%\STATE // generate a set of equations based on LBP $\varepsilon^{'}$ by mutating their arithmetic operators $A = \{+ ,  -, *, /\}$
			%\STATE // $op$ is the number of arithmetic operators \STATE // in the equation
			%\STATE $\mathcal{P}^{'}_{k} \leftarrow$ \textit{arithmetic-operators-random($A, op$)}	
			\STATE // mutate each equation $\epsilon \subseteq \varepsilon$ by $A$ 
			\STATE  $\{\epsilon^{'}_k \} \leftarrow$ \textit{(1+1)-CMA-ES ($\epsilon_{k}$, A)} 
			\STATE $\mathcal{P}^{'}_{k} \leftarrow$ \textit{Texture-BGS($\epsilon^{'}_{k}, X)$}
			\STATE \textit{lbp-scores}$\leftarrow$
			\{$\epsilon^{'}_{k}$, $\mathcal{P}^{'}_{k}\} $
			\ENDFOR	
			\STATE // Select the best equation by its accuracy
			\STATE  \{$\mathcal{P}$, $\varepsilon^{'} \} $ = $\arg\max(\textit{lbp-score})$
			\STATE \textbf{Output:}  Best equation $\epsilon^{'}$
		\end{algorithmic}
	\end{algorithm}
	
	\section{Proposed Method}
	\label{proposed_method}
	%For the background subtraction task, diversity models are initially learned for each pixel contained in the first $N$ images, say training sequences $\varkappa =\{ \varkappa_1 , \varkappa_2 , ..., \varkappa_N \}$ where each $\varkappa_j (j = 1, ..., N )$ is a certain pixel over time $N$ described by $ \varepsilon^{'}$ features based on LBP. 
	%For the background subtraction task, each pixel is modeled as a group of weighted LBP histograms $m =\{ m_1 , m_2 , ..., m_N \}$ from a video sequence, where $N$ is defined by user. 
	For the background subtraction task, a group of weighted LBP histograms are initially learned for each pixel contained in $N$ images, say video sequences $x =\{x_1 , x_2 , ..., x_N \}$ where each $x_j (j = 1, ..., N )$ is a certain pixel over time. Let a pixel at a certain location, considered as the center pixel $c= ({x}_c, {y}_c)$ of a local neighborhood composed of $P$ equally spaced pixels on a circle of radius $R$. The LBP descriptor applied to $c$ can be expressed as:
	
	\begin{equation}
	\label{eq:0}
	LBP_P,_R(x_c,y_c)= \sum_{p=0}^{P-1} s\left(g_p - g_c\right)2^p
	\end{equation}
	where $g_c$ is the gray value of the center pixel $c$ and $g_p$ is the gray value of each neighboring pixel, and $s$ is a thresholding function defined as:
	
	\begin{equation}
	\label{eq:0.1}
	s(x) = \begin{cases}
	1 & \text{if $x \geq 0$}\\
	0 & \text{otherwise}
	\end{cases}
	\end{equation}
	
	From (\ref{eq:0}), it is easy to show that the number of binary terms to be summed is $\sum_{i=0}^{P-1}2^i=2^P-1$, so that the
	length of the resulting histogram (including the bin 0 location) is $2^P$. 
	According to the authors\cite{heikkila:2006}, a limitation of the LBP original equation is that it does not work very robustly on flat images where the gray values of the neighboring pixels are very close to the value of the center pixel. In order to make the LBP equations more robust, as in \cite{heikkila:2006} our VAE (see section \ref{sec:generate_lbp} for further details) is able to generate equations with the $a$ term modifying the thresholding scheme of the descriptor by adding $a$ term to $s\left(g_p - g_c\right)$ in (\ref{eq:0}).

	\subsection{Generate Multiple Equations Based on LBP}
	\label{sec:generate_lbp}
	Considered a search space containing a variety of possible equations based on LBP given by $\varepsilon = \{\epsilon_1,  \epsilon_2, \dots,  \epsilon_K \}$, where $K$ is the user-defined number of equations that can be generated.  Instead of hand-designed search space, we opted to design it through a Variational Autoencoder (VAE). It is a powerful neural generative architecture capable of learning an encoder and a decoder that leads to data reconstruction and the ability to generate new samples from the input data. Given a small set of input equations $l$ the VAE maps the $l$ onto a latent space with a probabilistic encoder $q_\phi(z \mid l)$ and reconstructs samples with a probabilistic decoder $p_\theta(l \mid z)$. A Gaussian encoder was used with the  following reparameterization trick:
	
	%They are based on the LBP modified by \cite{heikkila:2006} for background modeling. The authors proposed to modify the threshold scheme of the original LBP equation, replacing the term $s(Z-C)$  with the $s(Z-C+a)$ where $C$ corresponds to the gray value of the center pixel of a local neighborhood, $X$ to the gray values of $P$ pixels equally spaced on a circle of radius R and $a$ is a small value.
	
	\begin{equation}
	\label{eq:1}
	\small
	q_\phi(z \mid l) =  \mathcal{N}(z \mid  \mu_{_\phi}(l), \sigma_{_\phi}(l)) = \mathcal{N}(\in \mid 0, I) \cdot \sigma_{_\phi}(l) + \mu_{_\phi}(l)
	\end{equation}
	
	Gaussian distribution $\mathcal{N}(0, I)$ is the most popular choice for a prior distribution $p{_\psi}(z)$ in the latent space. The Kullback-Leibler divergence, or simply, the $\mathcal{KL}$ divergence is a similarity measure  commonly used to calculate difference between two probability distributions. To ensure that $q(z \mid l)$ is similar to $p(l \mid z)$, we need to minimize the $\mathcal{KL}$ divergence between the two probability distributions. 
	
	\begin{equation}
	\label{eq:2}
	\small
	\min \mathcal{KL}(q_{_\phi} (z \mid l) \| p_{\theta} (z \mid l)) 
	\end{equation}
	
	We can minimize the above expression by maximizing the following:
	
	\begin{equation}
	\label{eq:3}
	\small
	L(\phi, \theta: l) =  E_{z \sim q_{_\phi} (z \mid l)}(\log ( p_{\theta} (l \mid z)) - \mathcal{KL}(q_{_\phi} (z \mid l) \| p (z)) 
	\end{equation}
	
	As we see in Eq.(\ref{eq:3}), the loss function for VAE consists of two terms, the \textbf{\textit{reconstruction term}} that penalizes error between the reconstruction of input back from the latent vector and the \textbf{\textit{divergence term}} that encourages the learned distribution $q_{_\phi} (z \mid l)$ to be similar to the true prior distribution $p(z)$, which we assume following a unit Gaussian distribution, for each dimension $j$ of the latent space. 
	
	%Search Strategy
	%Search Strategy by (1+1)-CMA-ES 
	% Covariance Matrix Adaptation Evolution Strategy
	
	\subsection{Search Strategy by (1+1)-CMA-ES}
	\label{sec:search_strategy}
	Given a set of equations $\varepsilon = \{\epsilon_1, \epsilon_2, \dots,  \epsilon_K \}$, a variant of Covariance Matrix Adaptation Evolution Strategy (CMA-ES)~\cite{igel:2007, hansen:2001} which is called  (1+1)-CMA-ES~\cite{igel:2006} looks for the best equation by mutating the arithmetic operators $(+ ,  -, *, /)$ of each equation $\varepsilon \subseteq E$ resulting in a new set of mutated equations given by $\varepsilon^{'} = \{\varepsilon^{'}_1, \varepsilon^{'}_2, \dots,  \varepsilon^{'}_K \}$. (1+1)-CMA-ES is an elitist algorithm based on an evolution strategy that operates on the Cholesky factors of the covariance matrix, reducing the computational complexity to $O(n^2)$ instead of $O(n^3)$.  Given fitness functions $f: \mathbb{R}^n\rightarrow\mathbb{R}, \Phi \mapsto f(\Phi)$ to be minimized. In (1+1)-CMA-ES strategy, at each update iteration a new offspring $\Phi_{offspring} \in \mathbb{R}^n$ (candidate solution) is generated from its parent $\Phi_{parent} \in \mathbb{R}^n$ (current solution), the worse solution is replaced by the better one for the next iteration. The success of the last mutation is determined as follows.
	
	\begin{equation}
	\label{eq:4}
	\gamma_{succ} = \begin{cases}
	1 & \text{if $f(\Phi_{offspring})  \leq  f(\Phi_{parent})$}\\
	0 & \text{otherwise}
	\end{cases}
	\end{equation}
	
	After sampling the new candidate solution, the step size is updated based on the success $\gamma_{succ}$ from a learning rate and a target success rate. The step size procedure and the update of the Cholesky factors are described in detail in ~\cite{igel:2006}.
	
	% (1+1)-CMA-ES is an elitist algorithm based on an evolution strategy with covariance matrix that uses a success rule for step size adaptation. Instead of calculating the covariance matrix, the (1+1)-CMA-ES, updates its Cholesky factors directly by the Theorem described in ~\cite{igel:2006} reducing the computational complexity to $O(n^2)$ instead of  $O(n^3)$.
	
	%Details on how the (1+1)-CMA-ES updates the step size and the covariance matrix can be found in ~\cite{igel:2006}.
	
	%Details on how the (1+1)-CMA-ES updates the step size and the covariance matrix can be found in ~\cite{igel:2006}.

	\subsection{Background Detection}
	\label{sec:background_detection}
	
	We estimate the performance of a set of mutated equations given by $\varepsilon^{'} = \{\epsilon^{'}_1, \epsilon^{'}_2, \dots, \epsilon^{'}_K \}$ by an efficient and fast non-parametric method for background subtraction algorithm called Texture BGS~\cite{heikkila:2004, heikkila:2006}. The $\epsilon^{'} \subseteq \varepsilon^{'}$ that presented the maximum accuracy
	$\{ \arg\max_{j \in 1, \ldots, K} \mathcal{P}_{j}(x)\}$ is chosen to deal with a particular challenge that is commonly encountered in complex scenes. The $\mathcal{P}$ represents a set of accuracy for each $\epsilon^{'} \subseteq \varepsilon^{'}$. In Texture BGS initially, a new LBP histogram computed from a new video sequence is compared with the current model histograms using the histogram intersection as the proximity measure represented by $\jmath$. If the threshold for the proximity measure is below for all model histograms, the model histogram with the lowest weight is replaced with the new histogram and it receives a low initial weight. In addition, if a model histogram close enough to the new histogram was found, the bins of this histogram and weights are updated as described in \cite{heikkila:2006}. Next, the histogram's persistence is used to determine whether the histogram models are more likely to belong to the background histograms or not. The persistence is directly associated with the histogram’s weight. As a last phase of the updating process, the model's histograms are ordered in descending order according to their weights, and the first histograms are select as background histograms.
	%\begin{equation}
	%\label{eq:7}
	%\omega_{1,t} +  ... + \omega_{B,t} > T_B, T_B \in [0, 1]
	%\end{equation}
	%where $T_B$ is the user-defined threshold and $\omega$ the  weight of the model histogram. In the foreground detection step, the histogram computed from a new video sequence is compared with the current background histograms using the same proximity measure as in the background model update step. 
	Comparing the histogram computed from a new video sequence to the current background histograms using the same proximity measure as in the background model update step for $x$ as follows:
	\begin{equation}
	\label{eq:8}
	H(x) = \begin{cases}
	1 & \text{if $\jmath < T_P$}\\
	0 & \text{otherwise}
	\end{cases}
	\end{equation}
	
	A pixel is classified as a background pixel if $H(x)$ = 0 for at least one background histogram.
	The proposed approach is summarized in Algorithm \ref{alg:structure_discovery}. Note that, we explain the LBP equation discovery procedure for a particular scene, but the procedure is identical for different scenes that present distinct challenges. 
	%A pixel is classified as a background pixel if the proximity is higher than the threshold $T_P$ for at least one background histogram.
	
	%it is a non-parametric approach that models each pixel as a group of adaptive LBP histograms that are calculated over a circular region around the pixel. Initially, the LBP histogram of the given pixel is calculated from the new video frame and compared to the current model histograms using an intersection measure of the histogram. Then, the histogram's persistence in the model is used to decide whether the histogram models the background or not. The higher the weight, the more likely it is a background histogram. At the end of the updating procedure, the model's histograms are ordered in descending order according to their weights, and the first histograms are select as background histograms. 

	%A complete description of all steps of the proposed framework is presented in Algorithm \ref{alg:structure_discovery}. 
	%\subsection{LBP Generation}
	%\subsection{Generate multiple LBP feature set}
	
	%Dropout layer
	\begin{table*}[!t]
		\centering
		\scalebox{0.72}{
			\begin{tabular}{| l | r  | r |} 
				\hline
				\textbf{Hyperparameters} & \textbf{Range Values} & \textbf{Best Values} \\[0.5ex] 
				\hline\hline
				\toprule
				\multicolumn{3}{c}{\textbf{VAE instance hyperparameters}}\\
				\hline
				ENC\_HIDDEN - Number of features in the hidden of the encoder  & \emph{choice}([125, 256, 512]) & 125  \\ 
				DEC\_HIDDEN - Number of features in the hidden of the decoder  & \emph{choice}([512, 800])  & 512  \\ 
				ENC\_LAYERS  - Number of recurrent layers of the encoder  & \emph{choice}([1, 2, 4, 6])  & 6   \\
				DEC\_LAYERS - Number of recurrent layers of the decoder  & \emph{choice}([1, 2, 4, 6])  & 1   \\
				ENC\_DROPOUT - Dropout rate  of the encoder  &  \emph{choice}([0.01, 0.02, 0.01, 0.1,  0.2]) & 0.1 \\
				DEC\_DROPOUT - Dropout rate of the decoder  &  \emph{choice}([0.01, 0.02, 0.01, 0.1,  0.2])  & 0.01   \\
				\toprule
				\multicolumn{3}{c}{\textbf{Training hyperparameters}}\\
				\hline
				
				N\_BATCH  -  Samples per batch to load & \emph{choice}([32, 64, 512])  & 32 \\  
				LEARNING\_RATE - Learning rate & \emph{choice}([0.001, 0.005])  & 0.005 \\  
				OPTIMIZER - Optimization algorithms &  \emph{choice}([Adam, Adadelta, RMSprop])   & RMSprop   \\  [1ex] 
				\hline
		\end{tabular}}
		\begin{tablenotes}
			\tiny
			\item   \textbf{${}^\star$ If (\textit{ENC\textunderscore LAYERS}) or (\textit{DEC\textunderscore LAYERS}) $>$ 1, becomes a bidirectional GRU otherwise unidirectional GRU.}
		\end{tablenotes}
		\caption{\small List of hyperparameters used in the VAE network in the training step. In the second column the range value that each of these hyperparameters can assume and finally the last column shows the combination of the best values for each hyperparameter. The set of the best hyperparameter values is used as the configuration of our VAE that will be responsible for generating our search space containing a set of equations based on LBP.}
		\label{table:1}
	\end{table*}
	
	\begin{figure}[t]
		\centering
		\includegraphics[width=0.43\textwidth]{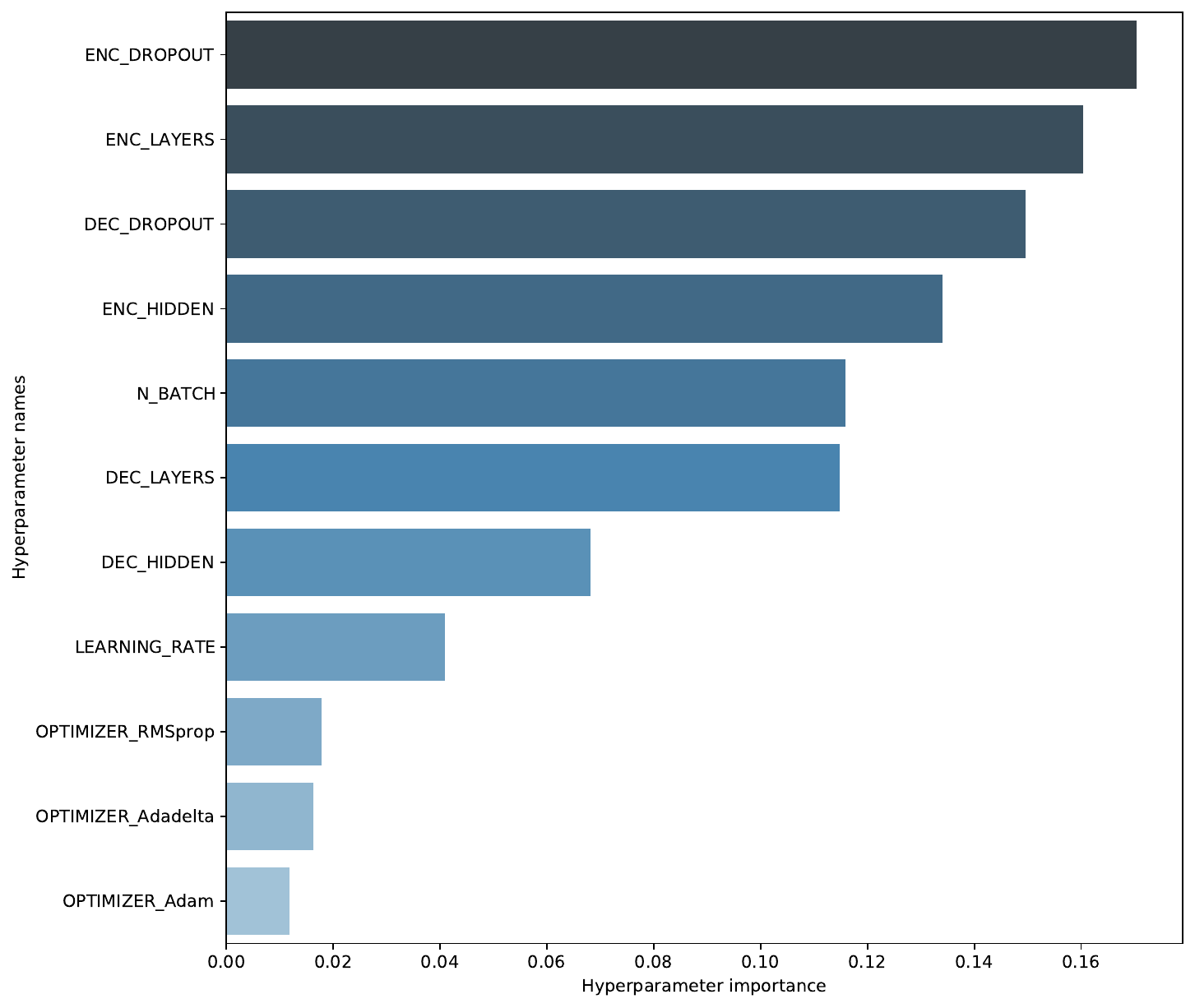}
		\caption{\small Hyperparameters Importance. The \textit{ENC\_DROPOUT, ENC\_LAYERS} and \textit{DEC\_DROPOUT} were the hyperparameters that had the greatest impact on training step.}
		\label{fig2:hyperimportance}
	\end{figure}
	%$s\left(g_p - g_c\right)$
	\begin{table*}[t]
		\centering
		\scalebox{0.72}{
			\begin{tabular}{ |p{2.6cm}|p{7.5cm}|p{9.2cm}|  }
				\hline
				\textbf{Scenes}     & \textbf{Challenges}  & \textbf{Best Discovered LBP Equation}\\
				\hline
				\hline
				\multirow{3}{7em}{\centering\em people in shade}  & Consists of pedestrian walking outdoor. The main challenges are: hard shadow cast on the ground by the walking persons and illumination changes.    &  \large(($g_p / g_c + a * g_c) - (g_p + g_c) - (g_p + g_c) + (g_p + g_c)) + a$ \\
				
				\hline
				\multirow{4}{7em}{\centering\em snow fall}  & Contains a traffic scene in a blizzard. The main challenges are: low-visibility winter storm conditions, snow accumulation, the dark tire tracks left in the snow.  &    \large($g_p - (g_p - g_c) * (g_p - g_c)) + a$ \\
				\hline
				% (Z-C)+(Z+C)*(Z+C)+(Z+C)/(Z/C)-(Z+C+C)/a 
				\multirow{3}{7em}{\centering\em canoe}  & Shows people in a boat with strong background motion. The main challenges are: outdoor scenes with strong background motion.  &    \large(($g_p /  g_c) - g_p) + a$  \\ %\large($g_p /  g_c - g_p + a$) \\
				\hline
				\multirow{3}{7em}{\centering\em bus station}  & Presents people waiting in a bus station. The main challenges are: hard shadow cast on the ground by the walking persons.  &   \large(($g_p /  g_c) - g_p) + a$ \\
				\hline 
				\multirow{3}{7em}{\centering\em skating}  & Shows people skating in the snow. The main challenges are: low-visibility winter storm conditions and snow accumulation.  &   \large(($g_p + g_c) * (a) - (g_p * g_c$)) \\
				\hline
				\multirow{3}{7em}{\centering\em fall}  & Show cars passing next to a fountain. The main challenges are: outdoor scenes with strong background motion.   &     \large(($g_p + g_c) /  (g_p - g_c) + a$) \\
				\hline
				
		\end{tabular}}
		\caption{\small List of the best equations for dealing with different challenges encountered in real-world scenarios. From the left to the right:  (a) different types of scenes, (b) with their main challenge descriptions, (c) the best LBP structure, and (d) the best LBP equation.}
		\label{tab:challenges} 
	\end{table*}
	
	\newcommand{\clarg}{25mm}
	
	\begin{table*}[t]
		\centering
		%	\scalebox{1.0}{
		\begin{tabular}{|l|l|c|c|c|}
			\hline
			{\bf Scenes}	&  {\bf Descriptors}&	{\bf Precision}	& {\bf Recall}	& {\bf F-score} \\
			\hline
			\hline
			\multirow{5}{\clarg}{\centering\em people in shade}		
			%	\multirow{4
			& Original LBP    &  0.6352  &  0.7321  &   0.6802 \\
			&  Modified LBP   &   0.7305 & 0.7725 &  0.7509 \\
			&  CS-LBP         &  0.4244  & {\bf0.8125}  &  0.5576	 \\
			& Proposed LBP & {\bf0.8163} & 0.8098 & {\bf0.8130} \\
			\hline
			\multirow{5}{\clarg}{\centering\em  snow fall}
			& Original LBP    &   {\bf0.9331} &   0.1425  &  0.2473 \\
			&  Modified LBP  & 0.8098 &   0.9119 &   0.8578 \\
			&  CS-LBP         &  0.5503 &  0.3150  &  0.4007	 \\
			& Proposed LBP         & 0.8592 &  {\bf0.8825} &  {\bf0.8707} \\    
			\hline
			\multirow{5}{\clarg}{\centering\em  canoe}
			& Original LBP        &  0.2295  &  0.2816  &  0.2529 \\
			&   Modified LBP      & 0.3410  &  0.4287  &  0.3798 \\
			&  CS-LBP             &  0.1866  &  0.3474  &  	0.2428 \\
			& Proposed LBP        &  {\bf0.8813} &  {\bf0.5428}  &   {\bf0.6719} \\
			\hline
			\multirow{5}{\clarg}{\centering\em  bus station}
			& Original LBP        & 0.3192 &  0.6127  &  0.4197 \\
			&  Modified LBP       &  0.6939 &  0.4022  &   0.5093 \\
			&  CS-LBP             &  0.0445  &  {\bf0.8083}  &  	0.0844 \\
			& Proposed LBP        & {\bf0.7120}  & 0.5124 & {\bf0.5959} \\ 
			\hline
			\multirow{5}{\clarg}{\centering\em  skating}	
			& Original LBP         &  0.6900 &  0.2704  & 0.3886  \\
			& Modified LBP        &  0.6839 &  {\bf0.8902}  &  0.7735 \\
			&  CS-LBP               &  0.1687 &  0.3527  &   0.2283 \\
			& Proposed LBP       & {\bf0.9178} & 0.7616  &  {\bf0.8324}\\
			\hline
			\multirow{5}{\clarg}{\centering\em  fall}	
			& Original LBP       &  0.6329 & 0.8778   &  0.7355 \\
			&  Modified LBP      &  {\bf0.8777} &  0.6328  &  0.7354 \\
			&  CS-LBP            &  0.3758 &  {\bf0.8783}  &  0.5264	 \\
			& Proposed LBP       & 0.8651  & 0.6701  &  {\bf0.7552}\\ 
			\hline
		\end{tabular}
		%}
		\caption{\small Performance using the CDnet 2014 dataset.}
		%	\normalsize
		%	\vspace{-5mm}
		\label{tab:performance} 
	\end{table*}
	
	\begin{figure*}[t]
		\includegraphics[width=\textwidth]{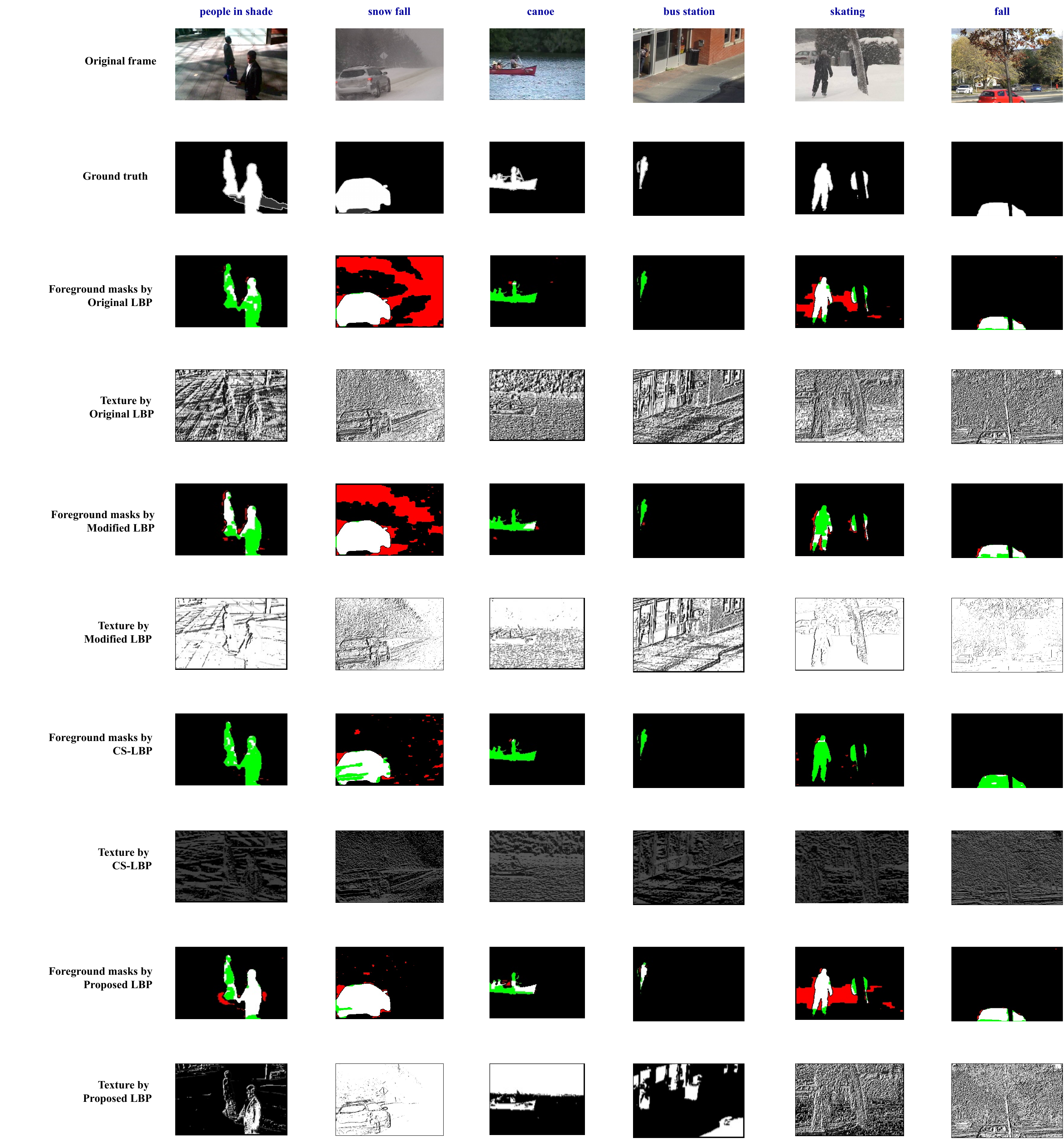}
		\caption{\small Background subtraction results using the CDnet 2014 dataset. From top to bottom: Original frame, Ground truth, Foreground masks by Original LBP, Texture by Original LBP, Foreground masks by Modified LBP, Texture by Modified LBP, Foreground masks by CS-LBP, Texture by CS-LBP, Foreground masks by LBP proposed and  Texture by LBP proposed. The true positives (TP) pixels are in white, the true negatives (TN) pixels are in black, the false positives (FP) pixels are in red, and the false negatives (FN) pixels are in green.}
		\centering
		\label{fig:visual_result3}
	\end{figure*}

	\section{Experimental Results}
	\label{sec:experimental_reults}
	In this section, we evaluated each component of our proposed method by conducting different experiments. First, we started by finding the set of optimal hyper parameters to train a VAE instance that generates a set of valid equations that are different from those within the training set. Our training set contains 80 equations based on LBP created manually by us. We also defined an evaluation measure called a number of \emph{Unseen} \& \emph{Valid Equations} (UVE), and our main objective in this first experiment was to find a configuration for the VAE instance that maximizes the UVE measure. A total of 300 VAE instances with different hyperparameter values were initially trained to generate a set of equations. The validity of each equation was verified by a regular expression. VAE models were based on the Gated Recurrent Unit (GRU)\cite{cho:2014}. It's an enhanced version of the standard recurrent neural network that aims to solve the vanishing gradient problem. We chose the GRU due to its simplicity and computation time. To perform the experiments optimally, we used an open-source platform called \emph{ProActive AI Orchestration (PAIO)} \footnote{https://doc.activeeon.com/latest/PAIO/PAIOUserGuide.html}. It allows researchers to easily automate machine learning (AutoML)\cite{hutter:2019, feurer:2019}, providing functions to automatically search for hyperparameters with parallel and distributed execution. We implemented the proposed method using PyTorch 1.0 and Python 3.7. The experiments were conducted on an NVIDIA GeForce RTX 2070 graphics card with 3.5 GHz Intel i7-3770K CPU that consists of 4 cores, 8 threads, and 16GB of RAM. Initially, we ran 150 interactions of two parallel instances of the VAE using the PAIO platform. We trained the models into 150 epochs using an early stopping mechanism. In Table \ref{table:1}, we tabulate the hyperparameters used in the VAE models in the training step. The first column contains the definitions of each hyperparameter. In the second column is the range value that each of these hyperparameters can assume, and finally the last column shows the combination of the best values for each hyperparameter. The set of the best hyperparameter values was used in the configuration of our VAE that will be responsible for generating our search space containing a set of valid and unique equations based on LBP. Note that the \emph{choice} represents a list of possible values for each hyperparameter. In Figure \ref{fig2:hyperimportance}, we show the importance of each hyperparameter when trying to maximize the UVE measure. As we can see the \textit{ENC\_DROPOUT, ENC\_LAYERS and DEC\_DROPOUT} were the hyperparameters that most impacted VAE training. This can be explained by the fact that the dropout layer is very important in VAE architectures, because they prevent all neurons in a layer from synchronously optimizing their weights, preventing them all from converging to the same goal, thus decorrelating the weights. Another hyperparameter that impacted the performance of our VAE was the number of recurrent layers in the encoder that compresses the original high-dimension input into the latent low-dimensional code. % https://stackoverflow.com/questions/50131402/variational-autoencoder-does-encoder-must-have-the-same-number-of-layers-as-the
	%  In \cite{heikkila:2006}, the authors suggest using a small value for $|a|$ in order to retain the discriminate power of the LBP descriptor. 
	A curious fact is that the number of encoder layers is bigger than the decoder. This was explained in \cite{chen:2017}, the authors showed if the decoder layer presents the same number of layers as the encoder, than there is a risk that VAE might completely ignore learning, therefore, keeping the $\mathcal{KL}$ loss to 0. Our VAE generated a total of 305 new equations. Some of the equations generated by our best VAE may present the $a$ term (see section \ref{proposed_method}). The (1+1)-CMA-ES modified each equation using ($4^{\eta}$ permutations), where $4$ represents  the four basic arithmetic operations in maths $[+ ,  -, *, /]$) and $\eta$ is the number of arithmetic operators presented in each equation. However, to prevent the generation of exponential mutations, we limited our experiment to $4^{5}= 1024$ mutations. In (1+1)-CMA-ES step, our main goal was to seek an equation for each scene that presents a better loss (1- F-score) than the hand-crafted equations compared in this paper, instead of conducting an exhaustive search. Note that this step of our approach can be fast if it finds the best equation at the beginning of the search. The search for the equations by (1+1)-CMA-ES was performed in a parallel and distributed manner.  In order to verify if the mutated equations are useful for dealing with different challenges found in real-world scenarios, we considered as our performance estimation strategy the Texture BGS approach (see subsection \ref{sec:background_detection} for more details) and a set of real-world scene sequences from CDnet 2014 dataset \cite{wang:2014}. We compared the new equations of the LBP descriptors discovered by our approach with three other texture descriptors among the reviewed ones, namely:
	
	\begin{itemize}
		\item{Original LBP \cite{ojala:2002}}
		\item{Modified LBP \cite{heikkila:2006}}
		\item{CS-LBP \cite{heikkila:2009}}
	\end{itemize}

	% and we downscale them to half the starting resolution
	We chose this one last descriptor because many recent hand-crafted LBP variants have been proposed based on their mathematical formulation \cite{bouwmans:2018}. The CS-LBP  generates compact binary patterns by working only with the center-symmetric pairs of pixels. For all descriptors, the neighborhood size is empirically selected so that $P = 8$, and $R = 1$. In addition, the thresholding value, $T$, from the CS-LBP was selected to $T = 0.01$ as in \cite{heikkila:2009}. We evaluated the performance of each descriptor in \emph{'people in shade', 'snow fall', 'canoe', 'bus station', 'skating'}, and \emph{'fall'} sequences. The description of each scene and its main challenges are presented in Table \ref{tab:challenges}. We limited the number of sequences for each scene to around $150$ frames choosing the sequences with a high degree of variability in terms of background changes, environmental conditions, hard shadows, suddenly start moving, etc. In addition, we down scaled the sequences to half the starting resolution due to the limitation of computational resources, since our approach can have high computation time. %In our experiments, the five first frames were enough to build a background model for each scene. 
	We present the visual results on individual frames from six different  scenes:  \emph{'people in shade'} (frame \#316), \emph{'snow fall'} (frame \#2758),  \emph{'canoe'} (frame \#904), \emph{'bus station'} (frame \#350), \emph{'skating'} (frame \#1884) and \emph{'fall'} (frame\#3987)   of the CDnet 2014 dataset. The best $a$ values were \emph{4.46, 8.03, 11.05, 8.28, 13.87, 0.67} for the Modified LBP and \emph{1.67, 57.97, 83.51, 56.10, 31.55, 3.32} for the proposed approach to \emph scenes {'people in shade',  'snow fall', 'canoe', 'bus station', 'skating' and 'fall'}, respectively.  The $a$ term was varied as $a=[10^{-2}, ..., 10^{2}]$. Note that the values of $a$ varied from small values to high values for both Modified LBP and proposed LBP's. Despite the authors in \cite{heikkila:2006} have suggested using a small value for $a$ for Modified LBP in order to retain the discriminate power of the LBP descriptor.
	%#\ref{fig:visual_result} \ref{fig:visual_result2}
	Figure \ref{fig:visual_result3} shows the foreground detection results using the Texture BGS method on the CDnet 2014 dataset. They were shown without any post-processing technique, except for the CS-LBP descriptor (see Figure \ref{fig:visual_result3}, line 8).  In order to increase the reader's understanding, we adjusted brightness and contrast for texture by CS-LBP. It is because the length of the resulting histogram of CS-LBP is more compact binary patterns $2^{4}$ than $2^{8}$ for the Original LBP, Modified LBP and LBP's proposed. The results obtained by the best equations discovered by the proposed method (see Table \ref{tab:challenges}) clearly appears to be less sensitive to background subtraction challenges and are able to detect moving objects with fewer false detection, especially in the \emph{'people in shade'} video that presents challenges such as hard shadow and intermittent shades, and in the \emph{'skating'} and \emph{'snow fall'} which are complex scenes with bad weather. Next, given the ground truth data, the accuracy of the foreground segmentation is measured using three classical measures:  recall,  precision, and  F-score. Table \ref{tab:performance} shows the proposed approach evaluated in the six scenes.  The best scores are in bold.  The proposed approach presented the best F-score for all scenes.  However, our approach encountered challenges in finding the best equation for scenes with dynamic background motion such as \emph{'canoe'} and \emph{'fall'}. Another challenging scene was \emph{'bus station'} which presents hard shadows. Nevertheless, the proposed LBP for \emph{'bus station'} scene (see F-score in Table \ref{tab:performance}) achieved an improvement of up to 3 times, overcoming the hand-crafted equations compared in this work. All the best equations discovered by our method, presented the $a$ term showing the importance of this term in image areas where the gray values of the neighboring pixels are very close to the center pixel one, e.g. sky, grass, etc.\cite{heikkila:2006}. We can see in Table \ref{tab:performance} that the original LBP and CS-LBP presented the worst performance for most scenes, such as \emph{'snow', 'fall', 'canoe', 'bus station'} and \emph{skating}. This can be explained by the fact that these two descriptors do not have the $a$ term in its equations. On the other hand, Modified LBP achieved a good F-score for most scenes, being behind only the descriptors discovered by the proposed method. The original LBP, Modified LBP, and Proposed LBP had a very close F-score for the \emph{'fall} scene. While CS-LBP performed worse in this scene. This can also be explained by the strategy used in the CS-LBP which the texture is obtained by comparing only center-symmetric pairs of pixels producing more compact binary patterns. The experiments also showed the scenes that presented similar challenges did not have the same best LBP equation. This can be noticed by the couples of scenes [\emph{'people in shade'} and \emph{'bus station'}], [\emph{'canoe'} and \emph{'fall']} and [\emph{'snow fall'} and \emph{'skating'}]  (see Table \ref{tab:challenges}). In contrast, scenes like \emph{'canoe'} [ dynamic background motion] and \emph{'bus station'} [hard and soft shadow and intermittent shades] had the same best equation. It shows us that each scene is unique and its dynamics also contribute to the choice of the best equation. Discovering a specific equation for a given scene manually, is a laborious task that requires an in-depth knowledge of the scene and a trial and error process by experts. Therefore, the approach presented in this paper can be very useful for discovering equations by machine computing while saving a lot of manual time.  The results presented in this work are a non-exhaustive search for the best equations due to our limited computational resources. The best equation search time depended on the scene type and the LBP equation generated by our VAE. Finally, we noticed that the most expensive step of our approach is in computing the LBP equations, taking up to $3$ seconds per frame. Future improvements can be the implementation of the LBPs to CUDA programming to accelerate and optimize the computing of our method.

	\section{Conclusion}
	\label{sec:conclusion}
	
	In this paper, we propose a novel approach able to discovering suitable LBP equations for background removal in videos. The main objective of this method is to reduce human time by automatically discovering LBP formulas, in the hope that this will eventually lead to the discovery of equations that we may never have thought of yet. Experimental results in video sequences show the potential of the proposed approach and its effectiveness in dealing with the main challenges encountered in real-world scenarios.

	%\clearpage
	
	%%%%%%%%% REFERENCES
	{\small
		\bibliographystyle{ieee_fullname}
		\bibliography{egbib}
	}
	
\end{document}